\documentclass[runningheads]{llncs}
\usepackage{graphicx}
\begin{document}
\title{Symbol Emergence and \\ The Solutions to Any Task}
\titlerunning{The Solutions to Any Task}
%
\author{Michael Timothy Bennett}
\authorrunning{M. T. Bennett}
%
\institute{School of Computing, Australian National University, Canberra, Australia\\ \email{michael.bennett@anu.edu.au}}
\maketitle              
\begin{abstract}
The following defines intent, an arbitrary task and its solutions, and then argues that an agent which always constructs what is called an Intensional Solution would qualify as artificial general intelligence. We then explain how natural language may emerge and be acquired by such an agent, conferring the ability to model the intent of other individuals labouring under similar compulsions, because an abstract symbol system and the solution to a task are one and the same.

\keywords{tasks \and symbol emergence \and artificial general intelligence.}
\end{abstract}
\section{Introduction}

We begin by briefly examining how intent may be defined and communicated. First, one may state intent generally and without context. For example, one may intend to acquire money. To make such a statement is to describe a goal \cite{b23}, and so we define it as that. Second there is contextual intent, a rationale. For example, if one observes a family member of an addict confiscating their drugs, one may infer that the family member intends to prevent the addict from overdosing. To do so is to assume the family member is compelled by an attachment to the addict. We imbue their behaviour with specific purpose by assuming the goal it serves in general, constructing a rationale.
Conversely, one may provide context for one's own decisions. It is here that the relationship between intent and causal reasoning becomes apparent; if an action was taken in service of a goal then in a sense that goal caused the action.
To state or infer a rationale, one must define the goal in service of which a decision was made, and a chain of causal relations \cite{b5,b6} indicating how that goal was to be served (whether successful or not).  
However, possession of a rationale is not the ability to communicate it. Humans represent and communicate in terms of loosely defined abstractions, tailored to express what is most important both to ourselves \textit{and} those to whom we are speaking. Human comprehension is not limitless. Absent such tailoring and simplifying abstractions, a rationale of even moderate complexity may be uninterpretable \cite{b7,evans_2021b}.
In a programming language meaning is exact, specified in the physical arrangement of transistors. Natural language is an emergent phenomenon \cite{b10} in which meaning is not limited to exact instruction \cite{b11}. Such a language must be interpreted, which suggests it is a means of encoding, transmitting and decoding more complex information. What is this complex information, what is the interpreter, and how would we build it? 

\paragraph{\bf Symbolic Abstraction:} Symbolic approaches to intelligence often adopt as their premise a Physical Symbol System \cite{newell_1990,nilsson_2007} grounded in hardware, yet actually implementing such a system is no trivial matter. This is known as The Symbol Grounding Problem \cite{harnad_1990}. 
Before continuing we must define what a symbol is. A computer scientist might define it as a dyadic relationship between a sign and an thing to which it refers (called a referent). This makes sense in the context of a programming language where an exact definition is necessary. We know how to implement such a thing using hardware, and so we'll call such dyadic symbol systems ``physically implementable languages".
However, dyadic symbols are not abstractions akin to symbols in natural language. Yes, ambiguity could be introduced in a dyadic symbol by linking a sign to a set of possible referents instead of one, and context could be captured by simply embedding contextual information in referents. Such symbols can and have been learned using existing machine learning methods \cite{b16,b17,b10}. As we will argue the problem lies in the construction of an abstract symbol system as a whole (not piece-wise), in what constitutes a symbol and the nature of messages communicated in natural language. 
Symbols determine about what one may reason, and so one symbol system may facilitate success in a given task better than another. We will describe how symbolic abstractions may be constructed to describe a specific task.
As to the nature of messages, Peircean semiosis \cite{b16} attempts to describe natural language on a conceptual level, defining a symbol as a triadic relationship in which the sign and referent are connected by an interpretant, which determines the effect upon the interpreter. What exactly an interpretant is and how it is implemented is part of what this paper seeks to clarify. Recall that in the example above we inferred the rationale of the hypothetical drug addict's family member's behaviour by assuming their goal. The family member's behaviour was a signal, and the goal served to decode that signal into a message; their rationale. We will argue that understanding natural language is not merely the result of clustering sensorimotor stimuli, but of imbuing stimuli with significance in terms of a goal. 
We could hardly claim to do all this by describing a goal using abstract symbols, and so it must be constructed in a physically implementable language. For simplicity of explanation we will describe a goal as a statement which has a binary truth value. Such a statement is true of a subset of possible the hardware (sensorimotor system) states, and false of others, however we see no reason goals with more degrees of truth would not suffice.
Hard coding such a goal is impractical, and so it must be learned by interacting with the world. We draw upon enactive cognition \cite{b14}, in which cognition is embodied, situated and extending into an environment. 
If a goal is to be learned then the question remains; which goal? The question is which is most plausible, most likely to generalise, and herein lies the connection to AGI.
Ockham's Razor is the notion that the simplest explanation is the most likely to be true. AIXI \cite{b27} is a theoretical artificial general intelligence which employs a formalisation of Ockham's Razor \cite{solomonoff_1964a,solomonoff_1964b} to decide which model of the world is most plausible. Solomonoff Induction formalises this notion by measuring the complexity of a program by it's Kolmogorov Complexity \cite{kolmogorov_1963}; the smallest self extracting archive which (in this case) reconstructs the agent's past experiences of the world. As a result, AIXI will learn the most accurate predictive models possible given what it has observed of the world. While such a model deals with programs rather than goal constraints, and is incomputable, it illustrates what the simple notion of Ockham's Razor is capable of; it not only bears out in anecdotal experience, but is of a deeper mathematical significance. To decide which goal is most plausible, we can follow a similar line of inquiry to formalise Ockham's Razor in terms of statements.
We draw upon preceding work for the formalisation of an arbitrary task and intent (using The Mirror Symbol Hypothesis) \cite{bennett_2021,b29,b7}, but differ in our characterisation of the solutions to an arbitrary task, how tasks can be subdivided or merged, and the relevance of AIXI. We have also abandoned notions such as perceptual symbols, introduced physically implementable languages, redefined abstract symbols and their relation to tasks and briefly addressed the emergence of normativity \cite{FitzPatrick_2021}.

\section{An Arbitrary Task}
In order to examine goals, we must define an arbitrary task. We do so by drawing upon boolean satisfiability problems to represent the task in terms of hardware states (not abstract symbols).
\begin{itemize}
    \item A finite set $\mathcal{X} = \{X_1, X_2, ..., X_n\}$ of binary variables.
    \item A set $\mathcal{Z}$ of every complete or partial assignment of values of the variables in $\mathcal{X}$, where
    \begin{itemize}
        \item an element $z \in \mathcal{Z}$ is an assignment of binary values $z_k$, which is $0$ or $1$, to some of the variables above, which we regard as a sequence $\langle X_i = z_i, X_j = z_j, ..., X_m = z_m \rangle$, representing a hardware or sensorimotor state. 
    \end{itemize}
    \item A set of goal states $\mathcal{G} = \{z \in \mathcal{Z} : C(z)\}$, where
    \begin{itemize}
        \item $C(z)$ means that $z$ satisfies, to some acceptable degree or with some acceptable probability, some arbitrary notion of a goal.
    \end{itemize}
    \item A set of states $\mathcal{S} = \{s \in \mathcal{Z} : V(s)\}$ of initial states in which a decision takes place, where
    \begin{itemize}
        \item $V(s)$ means that there exists $g \in \mathcal{G}$ such that $s$ is a subsequence of $g$, in other words for each state in $\mathcal{S}$ there exists an acceptable, goal satisfying supersequence in $\mathcal{G}$.
    \end{itemize}
\end{itemize}
The process by which a decision is evaluated is as follows:
\begin{enumerate}
    \item The agent is in state $s \in \mathcal{S}$.
    \item The agent selects a state $r \in \mathcal{Z}$ such that $s$ is a subsequence of $r$ and writes it to memory.
    \item If $r \in \mathcal{G}$, then the agent will have succeeded at the task to an acceptable degree or with some acceptable probability.
\end{enumerate}
For the sake of brevity, we will call $s$ which is given a situation (which may include memories of past experience), and $r$ which is selected a response. The distinction between situation and response is that a situation provides context for a decision, while a response is the result of one. We make no comment about any state following $r$, of which $r$ may not be a subsequence. The point is to model a decision, not a chain of events. A response may describe anything, from complex plans to simple instructions for actuators or memory read/write operations. 
Context alone does not tell us which response $r \in \mathcal{G}$ is correct for situation $s \in \mathcal{S}$; we need a goal constraint, a statement whose truth value determines correctness. Such a statement is necessary to determine whether any response given a situation is correct, and sufficient to reconstruct $\mathcal{G}$ from $\mathcal{S}$ (it need not reconstruct $\mathcal{S}$). There may be many statements which meet these criteria, but not all of them are what we might intuitively label \textit{the} goal. We will name this set of statements the domain of solutions to a task, and each such statement a solution.

\paragraph{\bf The Solutions to Any Task:}
\label{solutions}
A solution can be written using any physically implementable language $\mathcal{L}$ such as the aforementioned arrangements of transistors. 
To distinguish between possible solutions, we draw upon the notion of intensional and extensional definitions to be found in the philosophy of language \cite{b19}. For example, the extensional definition of the game chess is the enumeration of every possible game of chess, while the intensional definition could be the rules of chess. 
However, any statement could be a rule. In what way are the constraints we call the rules of chess any different from simply listing every legitimate game of chess? There is more than one set of rules which amount to the game chess. What we choose to call the rules of  any given game intuitively tend to be the weakest, most general individual rules necessary to verify whether any given example of a game is legal, and sufficient to abduct every possible legal game. Conversely, enumerating all valid games is just a means of  describing chess in terms of the strongest, most specific rules possible. The rules of chess describe the task “how to play chess”. However, there’s no reason we can’t extend the notion of rules from merely “how to play game $t$” to any arbitrary  task, such as “how to play game $t$ such that your chance of winning is maximised”. 
To reiterate, a rule is just a statement written in a physically implementable langauge. A statement is a solution if it is necessary and sufficient to reconstruct $\mathcal{G}$ from $\mathcal{S}$.
For every statement, there exists a set of hardware states of which it is true. The greater the cardinality of this set, the weaker the statement. To say one statement is weaker than another is to say it is true of more hardware states.
Given a physically implementable language of any practical use, there will exist connectives which can join two or more statements to form one stronger statement (for example ``and"), and connectives which can join two or more statements to form one weaker statement (for example ``or"). In either case, the resulting statement will be longer. Just as statements can be joined, a statement can be split into shorter separate statements by deleting a connective. The splitting of statements could continue until only atomic statements remain. At every split we could measure the weakness of the resulting statements, then the weakness of statements that result from the dissection of those and so on, to measure the overall statement by the weakness of its constituent parts. 
How specifically to go about measuring this is a matter for a much longer and more technical paper, but for now this suffices to illustrate how two solutions might be equivalent, but formed from very different constituent parts. As we are concerned with finding the most plausible statement, we'll consider two extremes of the domain of solutions:
\begin{enumerate}
    \item {\bf The Extensional Solution} to the task, formed from the strongest, most specific rules necessary and sufficient to abduct $\mathcal{G}$ from $\mathcal{S}$:
    \begin{enumerate}
        \item This is a statement $D$ enumerating every member of $\mathcal{G}$ as a long disjunction ``a or b or c or ... ". It is the longest solution possible without introducing redundant statements.
        \item It stipulates exactly each correct response for every situation, with no generalisation. It does not state what responses share in common.
        \item There is one and only one Extensional Solution given $\mathcal{L}$ and a task.
    \end{enumerate}
    \item {\bf The Intensional Solution} to the task, formed from the weakest, least specific rules necessary and sufficient to abduct $\mathcal{G}$ from $\mathcal{S}$:
    \begin{enumerate}
        \item This is a statement $C$ stipulating what the largest possible subsets of $\mathcal{G}$ share in common. 
        \item It stipulates what is necessary to verify the correctness of any response given a situation, but need not state the responses themselves.
        \item Its form depends upon $\mathcal{L}$, the physically implementable language employed, but if it were written in propositional logic in disjunctive normal form, then an Intensional Solution would be a disjunction of the shortest possible conjunctions necessary and sufficient to reconstruct $\mathcal{G}$ from $\mathcal{S}$.
        \item It adheres to Ockham's Razor in that it need not assert anything not strictly necessary to verify correctness (the rules it describes are together no stronger than is absolutely necessary).
        \item The above guarantees that any merely correlated variables will be eliminated from consideration, leaving only those relations most likely to be causal. 
        \item Just as there may be many functions which can interpolate a set of points, there may be many Intensional Solutions to a task given $\mathcal{L}$ (it may not be unique).
    \end{enumerate}
\end{enumerate}

\paragraph{\bf Intent Revisited:}
Earlier we defined intent as a goal, and it is certainly not the case that humans prefer to describe goals by enumerating every example of success. We try to describe what successes of a certain type share in common, we generalise. 
We will subsequently name an agent that always constructs Intensional Solutions an intentional agent, and one that always constructs Extensional Solutions a mimic. 

\paragraph{\bf Relationship to Ockham's Razor and AIXI:}
Given a task and an appropriate choice of physically implementable language, if we bundle each solution with a SAT solver and $\mathcal{S}$, then for each solution we have a self extracting archive that reconstructs $\mathcal{G}$. We will name this a solution archive.
The length of such a self extracting archive varies only with the length of the solution it employs ($\mathcal{S}$ being uncompressed, and the SAT solver being the same for all solutions). For each task there exists a unique value given $\mathcal{L}$; the length of the smallest solution archive. 
Note that this is not the Kolmogorov Complexity of $\mathcal{G}$, because we are not allowing $\mathcal{S}$ to be compressed, and we are only considering SAT solvers as decoders.
However, we can interpret Ockham's Razor as stating that an explanation should not assert anything more specific than absolutely necessary \cite{baker_2016}. A stronger statement is one that asserts more than a weaker statement in the sense that it is false in more hardware states than the weaker statement. This is what is important about Ockham's Razor, why it works; it minimises the possibility that the resulting statement is false. By this definition any Intensional Solution, not merely the shortest, should be sufficient to guarantee the most accurate prediction of goal satisfying responses possible. 
In contrast, the Extensional Solution is made up of statements each of which is false in all hardware states but one (because each one describes a unique $g \in \mathcal{G}$), minimising the plausibility of the solution's constituent parts.
Among the shortest solution archives there would exist an Intensional Solution, because a longer statement using a strengthening connective is less plausible than a shorter one, and at least one Intensional Solution must employ no more weakening connectives than strictly necessary (minimising length without losing necessary information). If we modified AIXI to consider only solution archives as models of the world, then it would be likely to find an Intensional Solution (because one must exist among the shortest solution archives it prefers). However, the shortest solution archive is not necessary to maximise predictive accuracy. If the reader accepts our characterisation of Intensional Solutions as being the most plausible according to Ockham's Razor, then any Intensional Solution will suffice.
Just as AIXI maximises reward across all computable environments, an intentional agent is one that attempts to maximise accuracy across all possible tasks. This is to say that in a specific task an agent possessed of a more specific inductive bias may outperform an intentional agent, but may not match an intentional agent in general. Just as lossless compression isolates causal relations \cite{b24}, so does an Intensional Solution. If we accept Chollet's \cite{b22} definition of intelligence as the ability to generalise, then these Intensional and Extensional Solutions represent the product of its extremes.
If, given a task, we choose a physically implementable language of such limited expressiveness that only a finite number of solutions exist, then an Intensional Solution is computable (by iterating through all possible solutions and comparing them). Of course, this only transfers the difficulty involved in constructing an AGI from the design of the AGI, to the design of the physically implementable language. 

\paragraph{\bf Learning a Solution:}
Learning typically relies upon an ostensive \cite{b21} definition; a small set of examples (hardware states, in this case) serving to illustrate what correctness is. An ostensive definition is defined as follows:
\begin{itemize}
    \item A set $\mathcal{G}_o \subset \mathcal{G}$ of goal satisfying states, which does not contain a supersequence of every member of $\mathcal{S}$.
    \item A set $\mathcal{S}_o = \{s \in \mathcal{S} : B(s)\}$ of situations (initial states) in which a decision takes place, where
    \begin{itemize}
        \item $B(s)$ means that there exists $g \in \mathcal{G}_o$ such that $s$ is a subsequence of $g$.
    \end{itemize}
\end{itemize}
Using this ostensive definition an agent could construct a solution which is necessary and sufficient to reconstruct $G_o$ from $S_o$. A solution is more general if, given $\mathcal{S}$, it implies goal satisfying responses to a larger subset of $\mathcal{G}$ than another solution. We'll examine $D_o$, the ostensive Extensional Solution constructed by a mimic, and $C_o$, an ostensive Intensional Solution constructed by an intentional agent. The mimic makes no attempt to generalise, and will eventually encounter a state $s \in \mathcal{S}$ where $s \not\in \mathcal{S}_o$ for which it knows no response. However, it may be possible for an intentional agent to construct $C_o = C$, meaning it learns the rules of the task as a whole. We say that an ostensive definition is sufficient if the ostensive Intensional Solutions it implies are necessary and sufficient to reconstruct $\mathcal{G}$ from $S$. If an ostensive definition is not sufficient, then the Intensional Solutions it implies are to a different task. An intentional agent would subsequently achieve optimal predictive accuracy for a task if given a sufficient ostensive definition. Among the possible solutions to a sufficient ostensive definition some will be more like an Intensional Solution than the extensional. An agent that learns such solutions will always generalise better than an agent that constructs solutions formed of stronger statements for the same reason an intentional agent generalises better than a mimic. If one can generalise more effectively from an ostensive definition, then one learns faster; every example added to the ostensive definition would convey a greater increase in predictive accuracy for the intentional agent than any agent that only constructs solutions closer to mimicry.

\paragraph{\bf Redemptive Qualities of a Mimic:}
The ability to generalise does not always serve a purpose. To illustrate, given a task to model a uniform distribution of goal satisfying responses, the Intensional and Extensional Solutions would both need to enumerate all goal satisfying states, and the entire contents of $\mathcal{G}$ would be required for a sufficient ostensive definition. While an intentional agent may be faced with computationally expensive abduction every time it needs to construct a response, a mimic's response would require minimal computation. It is akin to rote learning or human intuition, the ability to form a correct response without understanding what makes it correct. Fitting a function to a set of points may have more in common with mimicry than intent, which would explain why commonly employed machine learning methods require so much data yet struggle to generalise \cite{floridi_2020}. Perhaps to combine generalisability with computational efficiency, the best approach is to seek both Extensional and Intensional Solutions, the former to construct a heuristic.

\paragraph{\bf Constructing an Ostensive Definition using Objective Functions:}
Biological organisms are not usually given an ostensive definition with which to construct a solution to a task. Instead, we are compelled by primitives of cognition such as hunger and pain. These are, for all practical purposes, objective functions. By selecting those responses which resulted in favourable reward, we seem to construct ostensive definitions which we can then reason about and decompose into rules, identifying what members of an ostensive definition share in common. 

\section{Natural Language}
\label{explainable}
The interpretant of an abstract symbol, as we defined it, must not only cluster sensorimotor information but imbue it with significance in terms of a goal. By this definition, an Intensional Solution is playing the role of an interpretant, determining the effect stimuli has upon the interpreter's decisions. All incoming stimuli relevant to success in a task can then be perceived as a signal the agent must interpret. The Intensional Solution together with a SAT solver acts as a decoder, to compose an internal representation of a referent (a response). Conversely, any response which results in the agent taking an observable action could be perceived as emitting a signal. A solution may be split into shorter statements, each one of which may be perceived in isolation as the interpretant of an abstract symbol. The boundaries of a symbol are therefore fluid, as in natural language, dependent upon context. A sign or referent is any stimuli to which the interpretant refers, categorising stimuli in the same way it clusters sensorimotor states. The Intensional Solution may therefore be seen as both an abstract symbol itself, and a learned symbol system constructed specifically to efficiently describe what is important in a task.

\paragraph{\bf Communication:} 
For a signal to facilitate communication it must be imbued with similar meaning by both sender and receiver, what is called a normative definition. We posit that normativity emerges when interacting agents labour under similar compulsions, similar tasks (where a task may be defined so broadly as to encompass the human condition). This is because, in order to construct compatible symbol systems as we have described, two agents must construct approximately the same Intensional Solution, so that stimuli is imbued with similar meaning by both. Such a solution must account for the existence of other agents operating under the similar compulsions (otherwise an agent would never be compelled to respond to a situation by transmitting a signal), and so it must be learned in an environment where such other agents are present. If said agents have any significant impact on each others' ability to satisfy their compulsions, then solutions will imbue the observable behaviour of other agents with meaning. If co-operation is advantageous, then repeated interaction will produce conventions that facilitate complex signalling. As described earlier, a solution may be constructed using an ostensive definition, which may be constructed using objective functions. The task with which an agent engages is then determined by these objective functions. We will conclude this paper with an illustrative example of an agent learning an existing normative definition by interacting with others. 
For now, we illustrate what might be involved in encoding and decoding signals. To decode:
\begin{enumerate}
    \item Construct an Intensional Solution or something akin to it.
    \item Observe another agent responding to a situation, and apply one's own Intensional Solution to construct a rationale for their behaviour.
\end{enumerate}
In doing so an agent interprets what a signal means, what immediate sub-goals are being pursued. A signal in this form is not limited to spoken words but any behaviour, as in normal human interaction. However, if one's Intensional Solution provides no valid rationale for the observed behaviour of another individual, then the other agent's solution may be different from one's own. In this case, one must hypothesise modified Intensional Solutions that do permit a rationale, and in doing so simulate the possible goals of a creature different from one's own self.
To meaningfully transmit is to affect change in the sensorimotor state of another individual. One may convey or request information pertaining to sub-goals in order to co-operate, or perhaps to deceive in order to obtain some competitive advantage. In any case, the intentional agent is treating the other individual as part of the environment to be affected, then choosing a response which it predicts will satisfy its goals to some acceptable degree or with some acceptable probability. To encode:
\begin{enumerate}
    \item Decode the behaviour of others, predict their immediate sub-goals.
    \item How might they respond to your potential responses? Choose the response that you predict will result in their responding favourably.
\end{enumerate}

\paragraph{\bf An Example:}
Assume we have an intentional agent compelled by objective functions akin to those of a dog. We situate this agent within a community of dogs, and give it a body akin to a dog. Each day a human owner rings a bell when food is available, and each day the agent observes the sound of the bell, then the running of the other dogs to the location where food is placed, and then the food itself. Just as a statement may be split into shorter statements, so can a task be subdivided. The construction of a solution to an ostensive definition of the subtask we might call ``satisfy hunger" will associate the bell, the behaviour of the other dogs and the sight of food all with the satisfaction of the hunger compulsion, and so imbue them with meaning. These become symbols in an emergent language. Now consider the subtask ``avoid pain". A human approaching holding a stick, or a larger dog growling, will all be associated with pain after a few bad experiences. These are messages that convey the hostile intent of those other individuals. Having learned that a growl conveys hostile intent and the prospect of pain, the agent may attempt to reproduce this behaviour in order to obtain food claimed by a smaller dog, combining the rules of two subtasks in order to satisfy the goal of one.

\paragraph{\bf A Final Remark:}
Perhaps the most significant thing left to be said, which by now we hope is obvious, is that the solution to a task specifies an abstract symbol, and the solution to any subtask of that task also specifies an abstract symbol, and so an abstract symbol and symbol system amount to the same thing.
The above is an argument, not experimental proof. In future work we plan to construct experiments to test this idea, as well as the theory of tasks upon which it is based. At the time of writing we have constructed an agent that learns Intensional Solutions to binary arithmetic and other trivial tasks. Results pertaining to a more meaningful benchmark, which required the specification of a more expressive physically implementable language, are forthcoming. Finally, it is interesting to note that, because an Intensional Solution may be learned from only positive examples, it facilitates construction of a one-class classifier.


\begin{thebibliography}{00}

    \bibitem{b5} Pearl, J.: Causality: Models, Reasoning and Inference. 2nd. USA: Cambridge University Press. (2009)
    \bibitem{b6} Pearl, J., Mackenzie, D.: The Book of Why: The New Science of Cause and Effect. 1st. USA: BasicBooks, Inc. (2018)
    \bibitem{floridi_2020} Floridi, L., Chiriatti, M.: GPT-3: Its Nature, Scope, Limits, and Consequences. In: Minds and Machines pp. 1--14 (2020)
    \bibitem{b16} Taniguchi, T. et al.: Symbol Emergence in Cognitive Developmental Systems: A Survey. In: IEEE Transactions on Cognitive and Developmental Systems 11(4), pp. 494--516 (2019)
    \bibitem{b10} Taniguchi, T. et al.: Symbol Emergence in Robotics: A Survey. In: Advanced Robotics 30(11-12), pp. 706--728 (2016)
    \bibitem{b11} Santoro, A. et al.: Symbolic Behaviour in  Artificial Intelligence. Deepmind. arXiv: 2102.03406 [cs.AI]. (2021)
    \bibitem{newell_1990} Newell, A.: Physical Symbol Systems. In: Cog. Sci., pp. 135--183 (1980)
    \bibitem{nilsson_2007} Nilsson, N. J.: The Physical Symbol System Hypothesis: Status and Prospects. In 50 Years of Artificial Intelligence. Springer. pp. 9--17 (2007)
    \bibitem{harnad_1990} Harnad, S.: The Symbol Grounding Problem. (1990)
    \bibitem{b14} Thompson, E.: Mind in Life. In: Biology, Phenomenology and the Sciences of Mind 18 (2007)
    \bibitem{b17} Ramesh, A. et al.: DALL-E: Creating Images From Text. Open AI.  https://openai.com/blog/dall-e/ (2021)
    \bibitem{b19} Ostertag, G.: Emily Elizabeth Constance Jones. In: The Stanford Encyclopedia of Philosophy. Ed. by Edward N. Zalta. Fall 2020. Metaphysics Research Lab, Stanford University, (2020)
    \bibitem{b21} Gupta, A.: Definitions. In: The Stanford Encyclopedia of  Philosophy. Ed. by Edward N. Zalta. Winter 2019. Metaphysics Research Lab, Stanford University, (2019)
    \bibitem{b23} Setiya, K.: Intention. In:The Stanford Encyclopedia of Philosophy. Ed. by Edward N. Zalta. Fall 2018. Metaphysics Research Lab, Stanford University. (2018)
    
    
    \bibitem{bennett_2021} Bennett, M. T.: The Solutions to Any Task. PhD Thesis Manuscript. (2021)
    \bibitem{b7} Bennett, M. T., Maruyama, Y.: Philosophical Specification of Empathetic Ethical Artificial Intelligence. To appear in: IEEE Transactions on Cognitive and Developmental Systems (2021)
    \bibitem{b29} Bennett, M. T., Maruyama, Y.: Intensional Artificial Intelligence: From Symbol Emergence to Explainable AI. Manuscript. (2021)
    
    
    \bibitem{b22} Chollet, F.: On the Measure of Intelligence. arXiv: 1911.01547[cs.AI] (2019)
    
    \bibitem{kolmogorov_1963} Kolmogorov, A. N.: On tables of random numbers. In: Sankhya: The Indian Journal of Statistics A, pp. 369--376 (1963)
    
    \bibitem{b24} Budhathoki, K., Vreeken, J.: Origo: Causal Inference by Compression. In: Knowledge and Information Systems 56(2), pp. 28--307 (2018)
    
    \bibitem{b27}
    Hutter, M.: Universal Artificial Intelligence: Sequential Decisions based on Algorithmic Probability. Springer. (2005)

    \bibitem{solomonoff_1964a} Solomonoff, R. J.: A formal theory of inductive inference. Part I. In: Information and Control 7(1), pp. 1--22 (1964)
    
    \bibitem{solomonoff_1964b} Solomonoff, R. J.: A formal theory of inductive inference. Part II. In: Information and Control 7(2), pp. 224--254 (1964)
    
    \bibitem{baker_2016} Baker, A.: Simplicity. Stanford Encyclopedia of Philosophy. (2016)
    
    \bibitem{evans_2021b} Evans, R., Bošnjak, M., Buesing, L., Ellis, K., Pfau, D., Kohli, P., Sergot, M.: Making Sense of Raw Input. In: Artificial Intelligence 299, (2021)
    
    \bibitem{FitzPatrick_2021} FitzPatrick, W.: Morality and Evolutionary Biology. Stanford Encyclopedia of Philosophy. (2021)
\end{thebibliography}
\end{document}